\documentclass[conference]{IEEEtran}
\IEEEoverridecommandlockouts
\usepackage{cite}
\usepackage{amsmath,amssymb,amsfonts}
\usepackage{algorithmic}
\usepackage{graphicx}
\usepackage{textcomp}
\usepackage{xcolor}
\usepackage{multirow}
\def\BibTeX{{\rm B\kern-.05em{\sc i\kern-.025em b}\kern-.08em
    T\kern-.1667em\lower.7ex\hbox{E}\kern-.125emX}}
\begin{document}

\title{Lightweight Mask R-CNN for Long-Range Wireless Power Transfer Systems
}

\author{\IEEEauthorblockN{Hao Li\IEEEauthorrefmark{1}, Aozhou Wu\IEEEauthorrefmark{1}, Wen Fang\IEEEauthorrefmark{1}, Qingqing Zhang\IEEEauthorrefmark{1}, Mingqing Liu\IEEEauthorrefmark{1}, Qingwen Liu\IEEEauthorrefmark{1}, Wei Chen\IEEEauthorrefmark{2}}

\IEEEauthorblockA{\IEEEauthorrefmark{1}Dept. of Computer Science and Technology, Tongji University, Shanghai, China,\\
	\IEEEauthorrefmark{2}Dept. of Electronic Engineering, Tsinghua University, Beijing, China.\\}
\IEEEauthorblockA{ Email: \{lihao1101, 1732968, wen.fang, anne, clare, qliu\}@tongji.edu.cn,\\
			wchen@tsinghua.edu.cn}}

\maketitle

\begin{abstract}
	
Resonant Beam Charging (RBC) is a wireless charging technology which supports multi-watt power transfer over meter-level distance. The features of safety, mobility and simultaneous charging capability enable RBC to charge multiple mobile devices safely at the same time. To detect the devices that need to be charged, a Mask R-CNN based dection model is proposed in previous work. However, considering the constraints of the RBC system, it's not easy to apply Mask R-CNN in lightweight hardware-embedded devices because of its heavy model and huge computation. Thus, we propose a machine learning detection approach which provides a lighter and faster model based on traditional Mask R-CNN. The proposed approach makes the object detection much easier to be transplanted on mobile devices and reduce the burden of hardware computation. By adjusting the structure of the backbone and the head part of Mask R-CNN, we reduce the average detection time from $1.02\mbox{s}$ per image to $0.6132\mbox{s}$, and reduce the model size from $245\mbox{MB}$ to $47.1\mbox{MB}$. The improved model is much more suitable for the application in the RBC system.

\end{abstract}
 

\section{Introduction}

The rapid development of IoT and machine learning technology puts higher demands on high bit rate, low power consumption and big quantity of data \cite{b1}. Meanwhile, fifth-generation (5G) achieves breakthroughs in high speed, large capacity and low latency. However, with the growing contradiction between IoT development and power supply bottleneck, a significant increasement in the terminal endurance becomes a new requirement for sixth-generation (6G). Thus, wireless power transfer technology (WPT) may be a new dimension for 6G development \cite{b2}\cite{b3}. Resonant beam charging (RBC) system is a WPT which provides meter-level distance, watt-level power wireless charging services for mobile devices like smartphones \cite{b4}. The RBC system consists of the spatially separated transmitter and receiver. The transmitter transmits invisible beam to the receiver, which can then convert the beam power into electrical power for charging.

\begin{figure}[htbp]
	\centering
	\includegraphics[scale=0.2]{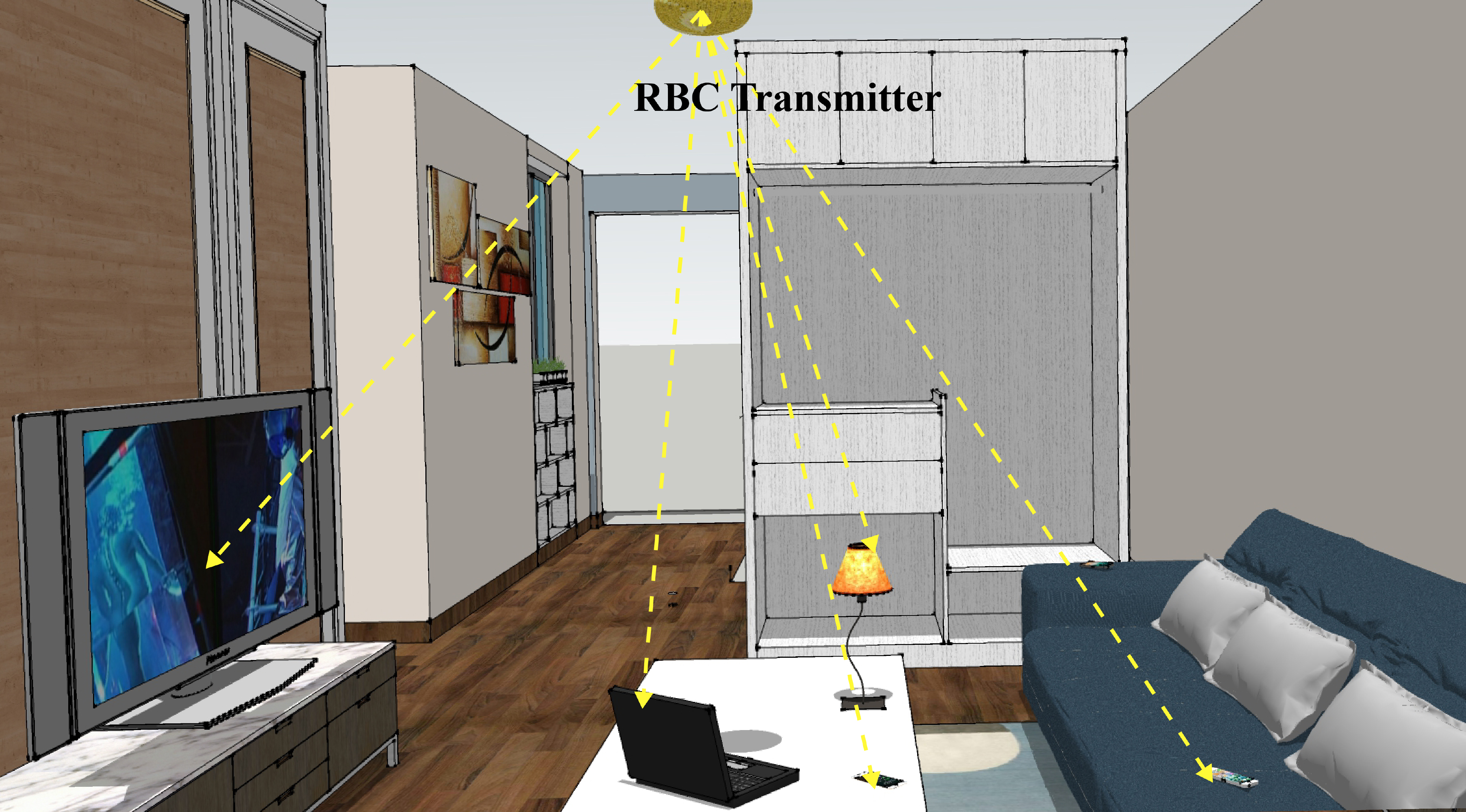}
	\caption{An indoor application scenario of the RBC system.}
	\label{introductionhao}
\end{figure}

Before transmitting beam power to a specific receiver, the transmitter should at first scan the whole area and determine the position of the receiver to be charged. Therefore, object detection method is needed to assist the transmitter to detect the receiver. Fig.~\ref{introductionhao} shows an indoor application scenario of the RBC system. After scanning and determining the charging targets, the RBC transmitter can charge multiple electronic devices simultaneously.

In \cite{b6}, the application of Mask R-CNN is applied to assist the RBC system in detecting the mobile devices to be charged \cite{b5}. However, the ResNet-50 based Mask R-CNN is not suitable for lightweight hardware-embedded devices because of its large parameter quantity and computation.

To reduce the hardware cost and the performance loss of object detection on embedded devices, the backbone of Mask R-CNN is changed from ResNet-50 to MobileNet V1 \cite{b11}\cite{b7}. Meanwhile, the ideas of Light-Head R-CNN are combined in the head part \cite{b8}. Since smartphones are the most widely used mobile devices and charging smartphones is the most common application of the RBC system, we choose smartphone as the detection object. The smartphone dataset built in \cite{b6} is used in the experiments.

\begin{table*}
	\caption{Computation and parameter quantity of ResNet-50 and MobileNet V1}
	\begin{center}
		\begin{tabular}{|c|c|c|c|c|}
			\hline
			\multirow{2}{*}{\textbf{Layer Name}} & \multicolumn{2}{|c|}{\textbf{ResNet-50}} & \multicolumn{2}{|c|}{\textbf{MobileNetV1}}\\
			\cline{2-5}
			& \textbf{Computation} & \textbf{Params} & \textbf{Computation} & \textbf{Params}\\
			\hline
			Conv 1 & $1.18,816,768$ & 9,664 & 41,746,432‬ & 3,648 \\
			Conv 2 & $672,358,400‬$ & 218,624 & 83,894,272‬‬ & 28,096‬ \\
			Conv 3 & $953,344,000‬$ & 1,226,752‬ & 80,325,504‬‬ & 105,344‬ \\
			Conv 4 & $1,389,273,088‬$ & 7,118,848‬ & 288,712,704‬ & 1,490,688‬ \\
			Conv 5 & $732,720,128‬$ & 14,987,264‬ & 77,923,328‬ & 1,601,024‬ \\
			\hline
			Total & 3.867GFLOPs & 23.561M & 0.573GFLOPs‬ & 3.229M‬ \\
			\hline
		\end{tabular}
		\label{tab1}
	\end{center}
\end{table*}

We perform comparison experiments on three frameworks: 
\begin{itemize}
\item ResNet-50 based Mask R-CNN.
\item MobileNet V1 based Mask R-CNN, in which we replace the backbone of Mask R-CNN from ResNet-50 to MobileNet V1.
\item MobileNet V1 based Mask R-CNN with light head, in which we replace the backbone with MobileNet V1 as well as adjust the head part based on Mask R-CNN.
\end{itemize}

The listed three frameworks are tested on the smartphone dataset. Our proposed approach shows faster detection speed and smaller model size, although at a compromise of slight accuracy reduce. After the adjustment, the average detection time is reduced from $1.02\mbox{s}$ per image to $0.61\mbox{s}$, and the model size is reduced by about $80\%$.

In section II of this paper, we will elaborate the adjustments of the Mask R-CNN framework. Then, the experimental analysis based on the smartphone dataset will be introduced. In section IV, we will show the improvement of charging efficiency and the traffic saving that the lightweight model brings to the RBC system. Finally, we will give conclusions and discuss open issues for further researches. 

\section{Lightweight adjustment of Mask R-CNN}

In this section, we give the implementation details for building the lightweight object detection model. 

Object detection frameworks based on convolutional neutral network (CNN) have two branches, one-stage framework and two-stage framework. For  two-stage frameworks, the features of image are extracted through the backbone, and then a region proposal networks (RPN) layer is followed to generate candidate regions. Finally, each proposal generated by RPN is classified and refined in the head part. Typical representatives of two-stage framework such as Faster R-CNN and Mask R-CNN, have higher accuracy but slower speed compared to one-stage framework \cite{b9}\cite{b5}. For one-stage framework, the RPN layer is not required but directly pinpoints and classifies objects of interest appearing in an image. The final results can be obtained after one detection, so the detection speed is faster. Typical representatives of one-stage framework are YOLO, SSD, etc \cite{b17}\cite{b18}. Although the detection speed of the one-stage framework is high, the performance is not yet comparable with two-stage framework. 

As two-stage frameworks are designed for multiple classes detection tasks (e.g., the COCO dataset contains 80 classes), a large backbone is used to extract finer features and a relatively heavier head part is used to classify objects. However, in the RBC system, where a small quantity of  mobile devices need to be identified, using a backbone like ResNet-50 to extract features is a bit overweight. Moreover, for lightweight hardware devices such as RBC transmitters, saving memory and improving detection speed are more meaningful for saving hardware costs and improving user experience.

As shown in Fig. 2, we make a lightweight improvement on the backbone and head parts based on Mask R-CNN, the backbone of Mask R-CNN is replaced with MobileNet V1, and the head part is changed to a lighter one. The adjustment greatly reduces the memory size and improve the detection speed with slightly sacrificing accuracy. 
 
\begin{figure*}[htbp]
	\centerline{\includegraphics[scale=0.45]{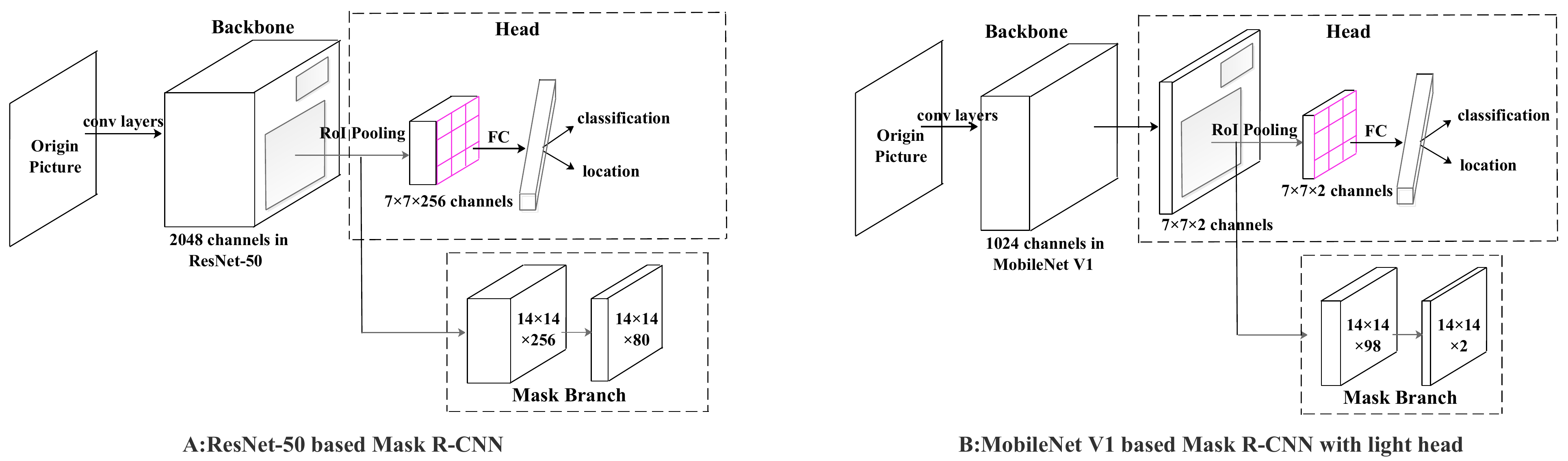}}
	\caption{Overview of the lightweight Mask R-CNN.}
	\label{fig}
\end{figure*}

\subsection{Adjustment in Body Architecture}

For two-stage frameworks like Faster R-CNN and Mask R-CNN, the method of extracting the feature map directly determines the detection speed and the model scale. The idea of MobileNet V1 that applying depthwise seperable convolution to reduce network capacity, is borrowed to reduce the parameter quantity and backbone computation \cite{b14}. According to \cite{b7}, MobileNet uses $3\times3$ depthwise separable convolutions which cost $8\sim9$ times less computation than standard convolutions at a compromise of little accuracy reduction.

Table I shows the difference of the floating point operations (FLOPs) and the quantities of parameters between ResNet-50 and MobileNet V1. Since MobileNet V1 uses depthwise seperable convolution to replace standard convolution, and the number of channels per convolutional layer is smaller than that of ResNet-50, both quantity of parameters and FLOPs of MobileNet V1 are much smaller than that of ResNet-50.

\begin{table}[htbp]
	\caption{Architecture of MobileNet V1 which is proposed in \cite{b7}}
	\begin{center}
		\begin{tabular}{|c|c|c|c|c|c|}
			\hline
			\textbf{Layer Name} & \textbf{Input} & \textbf{Block Operator}& \textbf{c} & \textbf{n} & \textbf{s} \\
			\hline
			Origin Picutre & 224*224*3 & Conv2d & 32 & 1 & 2 \\
			\hline
			& 112*112*32 & DepthWiseConv & 64 & 1 & 1 \\
			Conv 1 & 112*112*64 & DepthwiseConv  & 128 & 1 & 2 \\
			\hline
			& 56*56*128 & DepthwiseConv & 128 & 1 & 1 \\
			Conv 2 & 56*56*128 & DepthwiseConv & 256 & 1 & 2 \\
			\hline
			& 28*28*256 & DepthwiseConv & 256 & 1 & 1 \\
			Conv 3 & 28*28*256 & DepthwiseConv & 512 & 1 & 2 \\
			\hline
			& 14*14*512 & DepthwiseConv & 512 & 5 & 1 \\
			Conv 4 & 14*14*512 & DepthwiseConv & 1024 & 1 & 2 \\
			\hline
			& 7*7*1024 & DepthwiseConv & 1024 & 1 & 1 \\
			Conv 5 & 7*7*1024 & - & - & - & - \\
			\hline
		\end{tabular}
		\label{tab1}
	\end{center}
\end{table}

The architecture used in experiments is shown in Table II, which is proposed in \cite{b7}. Each line describes a sequence of one or more identical (modulo stride) layers, which will repeat $n$ times. All layers in the same sequence have the same number $c$ of output channels. The first layer of each sequence has a stride $s$ and the others use stride $1$. All spatial convolutions use $3\times3$ kernals. Layers $Conv 1\sim5$ are used in RPN network to get reigon of interests (RoIs) \cite{b12}. 

\subsection{Adjustment in Head Architecture} 

Classic two-stage framworks such as Faster R-CNN and Mask R-CNN, use two heavy fully connected (FC) layers for proposal prediction \cite{b15}. Moreover, each RoI needs to be calculated separately, which is time-costing. In addition, the number of feature channels after RoI pooling is large, which makes the two FC layers consume a lot of memory and potentially affects the computational speed.

To reduce the computation and the memory consumption in the head architecture, we borrow the idea of Light-Head R-CNN \cite{b8}. Light-Head R-CNN is presented on the basis of R-FCN, which compresses the after-pooling feature maps of R-FCN to $10\times P\times P$ (P is the followed pooling size) \cite{b10}. The process is equivalent to pressing more than $3900$ channels to $490$ channels when $P$ equals $7$, and a FC layer is added to output the final classification result. In this paper, we generate feature maps with small number of channels ($2\times 7\times7$ is used in the experiments) and only keep one single FC layer after the pooling layer. This adjustment greatly reduces the computational cost in the head part and the memory requirements of the detection system. In addition, we reserve the mask branch proposed in Mask R-CNN to get higher precision accuracy.

\section{Experiments}

In this section, we introduce the detailed training process of three different networks on the smartphone dataset. In the subsection B, the experimental comparison results are shown to demonstrate that the modified model is faster and lighter.

\subsection{Model Training}

We train and evaluate the model with Tensorflow and Open Source TensorFlow Object Detection API. Since the smartphone dataset is small, pretrained network parameters are needed for training. We train the MobileNet V1 based Mask R-CNN with the COCO 2017 dataset\cite{b13}, and load the pre-trained model on the imagenet to assist in training \cite{b16}. The NVIDIA GeForce GTX 1070 6G graphics card is used for training. Fig. 3 shows the detection sample on the trained model. The model size of the ResNet-50 based Mask R-CNN is $245M$, while the new one with MobileNet V1 is $93M$, which reflects the adjusted model is smaller and easier to be transplanted on embedded devices.

\begin{figure}[htbp]
	\centerline
	{\includegraphics[scale=0.37]{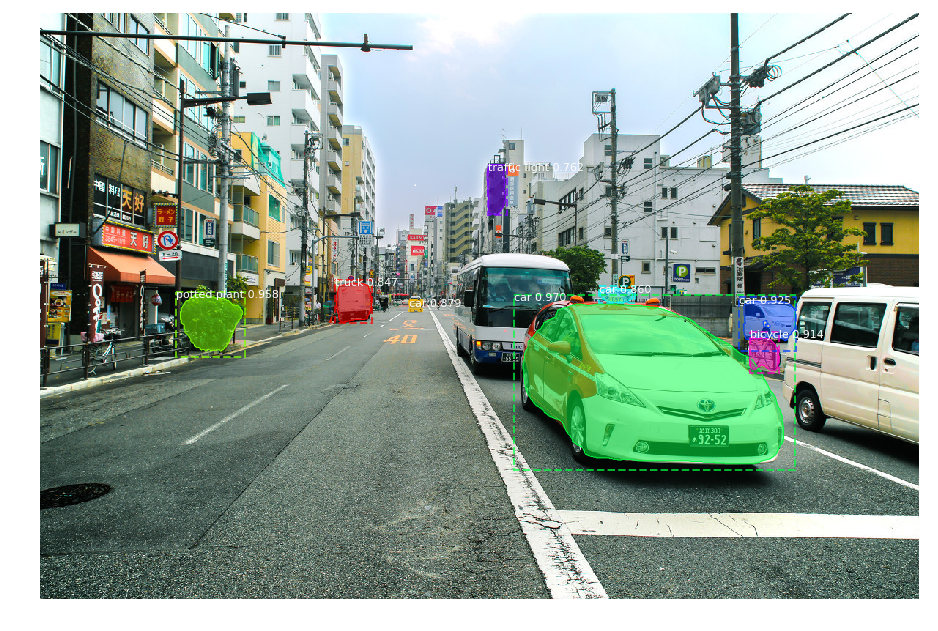}}
	\caption{Detection sample on the MobileNetV1 based Mask R-CNN framework}
	\label{fig}
\end{figure}

Then, we adjust the head architecture and load the pretrained model with excluding irrelevant layers. The smartphone dataset built in \cite{b6} is used to train our model. The entire dataset contains 3,200 images taken from classrooms, bedrooms, laboratories and other indoor scenarios. The entire dataset is divided into three parts, including training set with 1600 images, development set with 800 images, and test set with 800 images.

The training process is divided into three steps. First, train the mask branch and the two branches in the head part because they are completely randomly initialized. Second, fine-tune layers from layer $Conv 4$ and up. Third, all layers are fine-tuned with a smaller value of learning rate. At the end of each epoch during the training process, the fitting results are observed to prevent over-fitting and under-fitting. A detection sample on the smartphone test set is shown in Fig. 4.

\begin{figure}[htbp]
	\centerline
	{\includegraphics[scale=0.57]{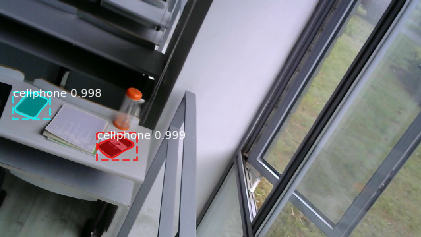}}
	\caption{Smartphone detection sample}
	\label{fig}
\end{figure}

\subsection{Experimental Analysis}

We test the detection speed (average detection time of 500 pictures), model sizes and mean average precisions (mAPs) of the three frameworks on the smartphone test set to evaluate the performance of our model. Table III shows the experimental results of the three different frameworks. It can be seen that the framework with MobileNet V1 and light head reduces the detection time by about $40\%$ and reduces the model size by about $80\%$ respectively compared with the ResNet-50 based Mask R-CNN.

\begin{table}[htbp]\label{key}
	\caption{Comparison among three frameworks based on the smartphone datasets}
	\begin{center}
		\begin{tabular}{|c|c|c|c|c|}
			\hline
			\multirow{2}{*}{\textbf{Backbone}} & \multirow{2}{*}{\textbf{Head}}& \textbf{Speed}& \textbf{Memory Size}& \textbf{mAP} \\
			
			 & & \textbf{(s/img)}& \textbf{(MB)}& \textbf{(\%)} \\
			\hline
			ResNet-50 & normal & 1.02 & 245 & {\bfseries 57.66} \\
			MobileNet V1 & normal & 0.6697 & 91.1 & 38.23 \\
			MobileNet V1 & light head & {\bfseries 0.6132} & {\bfseries 47.1} & 36.55 \\
			\hline
		\end{tabular}
		\label{tab1}
	\end{center}
\end{table}

In the experiments, average precision (AP) is used to evaluate the accuracy of our model. AP of the model is tested at 10 IoU thresholds (from 0.50 to 0.95 step by 0.05) and mAP is shown in Fig. 5. It is shown that AP is relatively high when IoU is around 0.55, and AP decreases linearly as the IoU increases. Although mAP of the lightweight model is lower than that of the other two frameworks, a slight decrease in accuracy with obvious improvement on detection speed and memory size is acceptable. 

To ensure safety under various complex circumstances, detection precision must be high enough in the applications such as auto driving. However, common application scenarios of the RBC system are indoor places. In these indoor scenarios, it is easy to achieve higher accuracy after the detection model being optimized. In addition, the RBC trainsmitter attempts to communicate with the mobile devices before charging. Even if the target is identified incorrectly, the RBC transmitter can perform charging operations by scanning all areas. It means that false detection will not cause any harm to the human body. Therefore, the lightweight object detection framework proposed in this paper is suitable for the RBC system which do not require strict accuracy. Moreover, it can be better applied to the RBC system to assist in object detection with the computation and time costs cut.

\begin{figure}[htbp]
	\centerline
	{\includegraphics[scale=0.60]{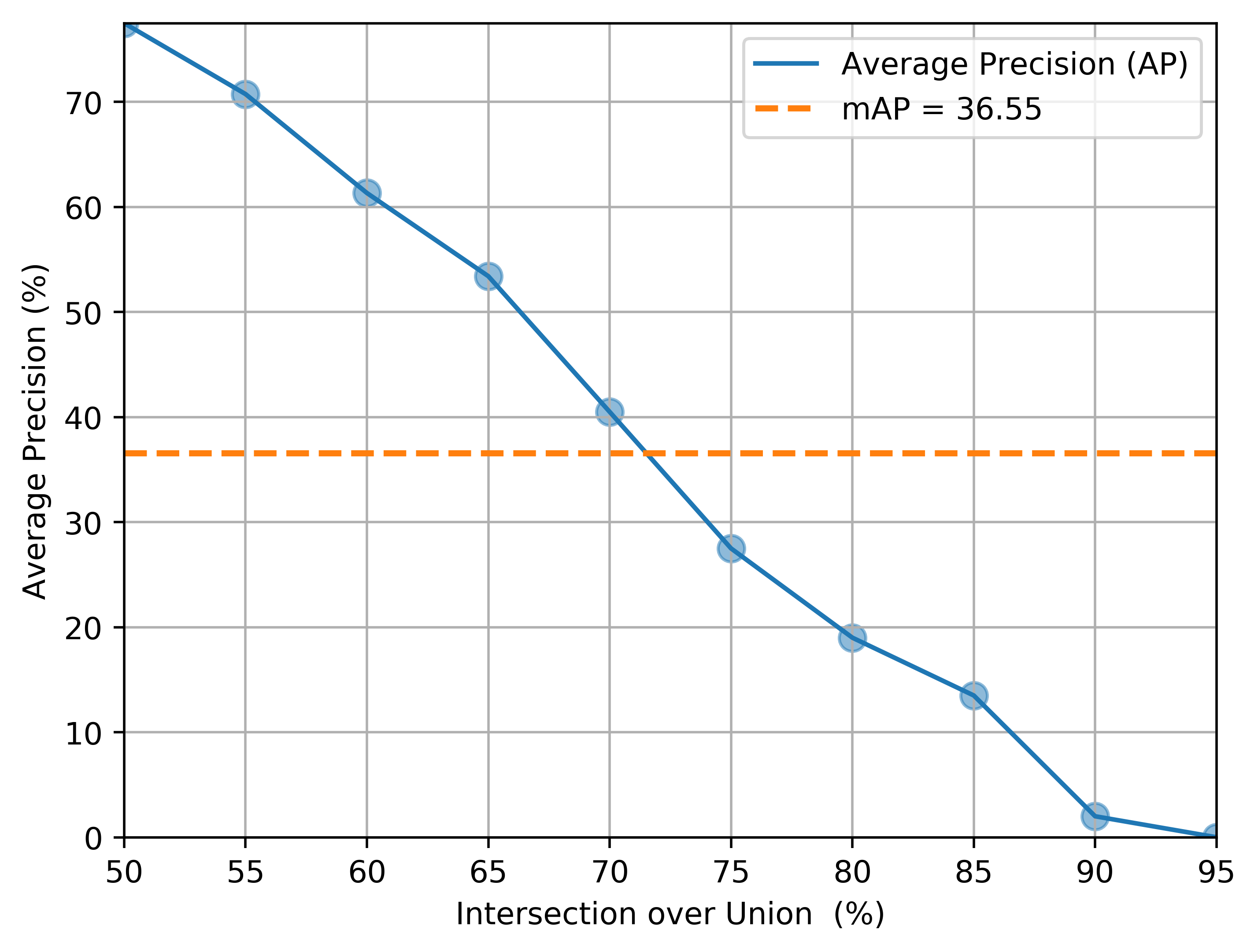}}
	\caption{Average precision vs. Intersection over union.}
	\label{fig}
\end{figure}

\section{Performance Analysis}

Applying object detection to the RBC system can effectively improve its working efficiency. It has been introduced in \cite{b6} that object detection methods can effectively help RBC transmitter to detect IoT devices. To apply object detection methods to the RBC system, the key is whether the embedded device can accommodate and run a large network. There are three factors limit the application of deep learning methods in embedded devices:
\begin{itemize}
\item The model parameters are large, occupying large memory.
\item The computational cost is high.
\item The corresponding software implementation of the embedded system is complex.
\end{itemize}

The model proposed in this paper alleviates the first two problems to a certain extent. For the third problem, there are some deep learning frameworks for embedded devices, such as DarkNet, Tiny-DNN, NCNN, MXNet, etc \cite{b19}.

There are three options for combining the RBC systems with object detection methods:
\begin{itemize}
\item The image is preprocessed locally, then uploaded to the server for deep processing, and the final result is processed locally.
\item Only the image collection is done locally, all processes are  completed on the server, and then the device receives the result.
\item Completely locally processed.
\end{itemize}

For complex deep neural networks such as ResNet-50 based Mask R-CNN, because of its huge computation and model size, it is not suitable for processing captured images locally. Only the first two solutions can be used to assist with processing. However, the lightweight framework proposed in this paper has only one-fifth the size of the original model, which greatly reduces the burden of hardware devices. It is feasible for RBC transmitters to process images locally when using the lightweight object detection framwork. The following subsections introduce the improvement of charging efficiency and traffic saving with the lightweight model.

\subsection{Detection Speed Increment}

Compared with the object detection processed on the cloud server, completely processed locally has obvious advantages in saving detection time. As shown in Fig. 6, we suppose the upload speed from the RBC transmitter to the cloud server is $S_{u}$, the download speed from the cloud server to the RBC transmitter is $S_{d}$ and the network delay is $T_{delay}$. The time for object detection model to process one image is $T_{detect}$ and the size of each image is $K_{image}$. The total time for processing one picture locally is $T_{local}$ and the total time for processing one picture on cloud is $T_{cloud}$. For $T_{local}$, image capture and object detection are done locally. After detection, the RBC trainsmitter send the collected information $K_{info}$ to the cloud server for verification, and finally perform charge action depends on the verification result $K_{result2}$. For $T_{cloud}$, because its object detection is done on cloud, the captured image needs to be uploaded to the cloud server for detection, and the cloud server returns the detection result $K_{result1}$. The rest action is the same as $T_{local}$. Then $T_{local}$ and $T_{cloud}$ can be depicted as:
\begin{align}
\label{Tlocal} T_{local}&= T_{detect} + \frac{K_{info}}{S_{u}} + \frac{K_{result2}}{S_{d}} + 2T_{delay} \\
\label{Tcloud} T_{cloud}&= T_{detect} + \frac{K_{image} + K_{info}}{S_{u}} + \frac{K_{result1} + K_{result2}}{S_{d}} \nonumber\\ 
&~~~ + 4T_{delay} 
\end{align}

We assume that ResNet-50 based Mask R-CNN is run on cloud, and the lightweight model proposed in this paper is run locally. Each image is 0.1MB. Ignoring $T_{delay}$, $K_{result1}$, $K_{info}$ and $K_{result2}$, the upload and download speed are all 1MB/s. The speed in Table II is used to substitute $T_{detect}$, and then we get $T_{local}/T_{cloud} = 0.5475$. That means processing images locally can reduce the process time by nearly half compared with that on the cloud server.

\begin{figure}[htbp]
	\centerline
	{\includegraphics[scale=0.45]{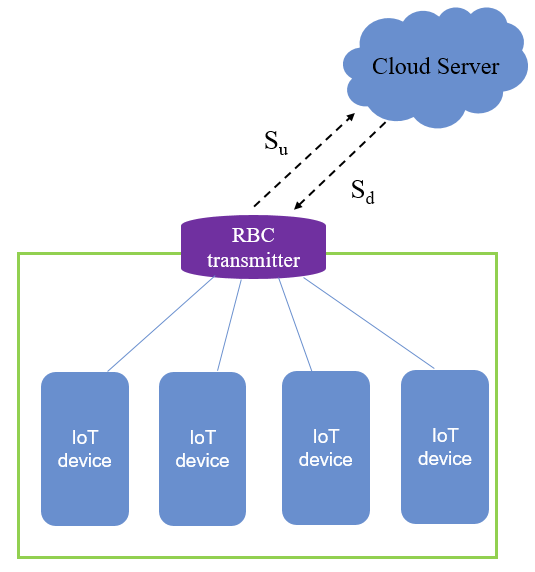}}
	\caption{An application scenario of the RBC system}
	\label{RBC_apply}
\end{figure} 

\subsection{Traffic Saving}

In the RBC system, adopting a local-running lightweight object detection model can save traffic. When object detection is performed locally, the RBC transmitter processes the collected picture itself and identify the devices that may need to be charged. Then the RBC transmitter send the collected information $K_{info}$ to the cloud for verification. Finally, based on the $K_{result2}$ returned from cloud , the transmitters determines whether to charge the devices or not. Thus, the traffic cost $C_{local}$ can be depicted as:
\begin{equation}
C_{local} = K_{info} + K_{result2} \label{clocal}
\end{equation}

If object detection is performed on cloud, the RBC trainsmitter needs to upload the collected image to get the processed information $K_{result1}$. Then, the RBC transmitter collect the device ID of RBC receiver depending on $K_{result1}$. The remaining steps are the same as that in the local process, the RBC trainsmitter determines whether to charge the receiver or not according to the verification result from the cloud server. In that way, the traffic cost per-detection on cloud $C_{cloud}$ can be depicted as:
\begin{equation}
C_{cloud} = K_{image} + K_{result1} + K_{result2} + K_{info} \label{ccloud}
\end{equation}

In practical applications, $K_{result1}$ and $K_{info}$ consume much less traffic compared to $K_{image}$. It can be seen from \eqref{clocal} and \eqref{ccloud} that,  for each time the detection is performed, locally processed can save traffic of more than one image compared to processed on cloud. Therefore, our proposed approach can greatly reduce the traffic cost in real applications.

\section{Conclusion}

In this paper, we present a lightweight two-stage object detection approach, which is easier to be transplanted on embedded devices to improve charging efficiency of the RBC system. We train the new model on a smartphone dataset. Experimental results show that the new model is much faster and lighter than the ResNet-50 based Mask R-CNN. Finally, we discuss the improvement in detection efficiency and traffic cost that the lightweight detection model bring to the RBC system. The improvement and application of the model proposed in this paper can be further studied from the following points:
\begin{itemize}
	\item The types of mobile devices and the applications identified currently are still relatively limited. We hope to identify more mobile devices in more application scenarios in the future.
	\item Many lightweight object detection methods can achieve fast detection speed and high accuracy nowadays. In future researches, we will study the combination of other lightweight frameworks with the RBC system.
	\item The current RBC applications are still based on theoretical researches. Object detection methods can be tested on real embedded devices for further research.
\end{itemize}

\end{document}